\documentclass{article}
\usepackage{spconf,amsmath,graphicx,hyperref}
\usepackage{booktabs}
\usepackage{amssymb}

\title{Diagnosis-Conditioned Spatial Gating and Decoder-Level Supervised Contrastive Learning for Radiology Report Generation}

\name{Md Mustafizur Rahman, Myl\`ene C. Q. Farias}
\address{Department of Computer Science, Texas State University, San Marcos, TX, USA}
\begin{document}
%
\maketitle

\begin{abstract}
Radiology report generators can be fluent yet inaccurate for specific findings. Diagnosis-driven prompting helps, but classifier outputs often do not explicitly condition the visual patch representations used as decoder cross-attention memory. Global channel gating can provide such conditioning, but it applies the same channel reweighting at every spatial location. We propose a Position-Aware Gate (PAG) that uses predicted finding states to spatially modulate patch features without region-level supervision. We also introduce a Decoder-level Supervised Contrastive Loss (DSCL) that structures decoder representations by treating studies sharing at least one ground-truth positive finding as positives. On MIMIC-CXR, PAG and DSCL individually improve clinical efficacy (CE) F1 from the Global+Align reference of 0.484 to 0.491 and 0.495, respectively. Combining both components further increases CE F1 to 0.502. Without fine-tuning on IU X-Ray, the combined model achieves 0.226 CE F1 compared with 0.211 reported by PromptMRG.
\end{abstract}


\section{Introduction}
\label{sec:intro}

\begin{figure}[t]
\centering
\fbox{%
\begin{minipage}[t]{0.97\columnwidth}
\begin{minipage}[t]{0.31\linewidth}
\vspace{0pt}
\raggedright
\includegraphics[width=0.95\linewidth]{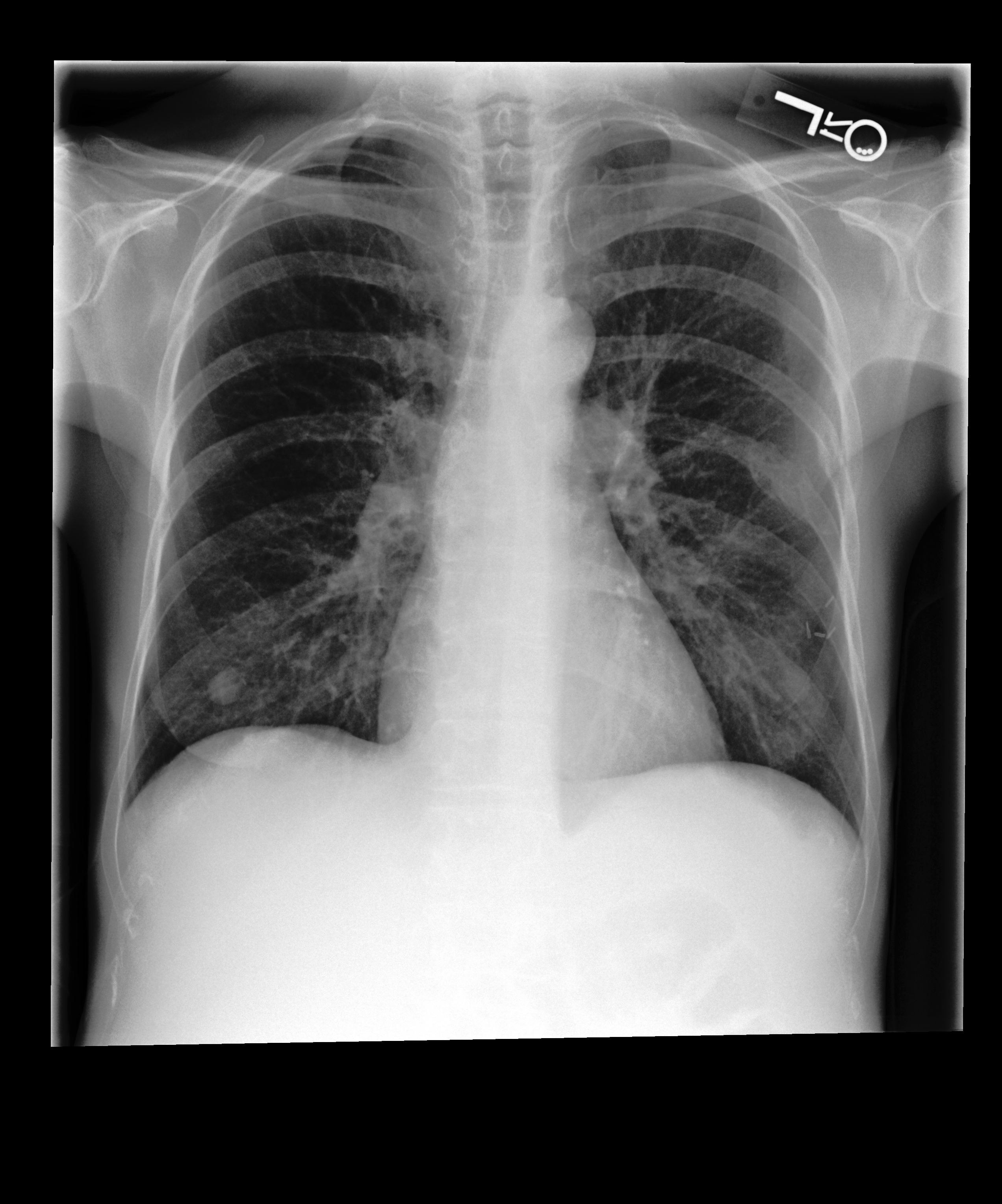}
\end{minipage}
\hspace{1mm}
\begin{minipage}[t]{0.66\linewidth}
\vspace{0pt}
\small
\textbf{Reference report:}

\vspace{2pt}
There is no focal consolidation, pleural effusion, or
pneumothorax. Bilateral nodular opacities that most likely
represent nipple shadows. The cardiomediastinal silhouette
is normal. Clips project over the left lung, potentially
within the breast. The imaged upper abdomen is unremarkable.
\end{minipage}
\end{minipage}%
}
\caption{Example chest X-ray and reference report from
MIMIC-CXR~\cite{johnson2019mimiccxr}.}
\label{fig:example_report}
\end{figure}

Radiology report generation (RRG) maps medical images to free-text reports. Figure~\ref{fig:example_report} shows a chest X-ray with its reference report. Although recent models produce fluent text, high natural language
generation (NLG) scores may not reflect clinical accuracy. Generated reports can omit abnormalities or contain factual inconsistencies~\cite{miura2021factual}, while lexical-overlap metrics may miss clinically important errors~\cite{yu2023evaluating}. Accordingly, RRG studies also report clinical efficacy (CE) metrics based on structured findings extracted using labelers such as CheXbert~\cite{smit2020chexbert}.

Prior work uses clinical or spatial guidance to improve report generation. RGRG~\cite{tanida2023rgrg} generates region-conditioned sentences using a detector trained with bounding-box supervision. PromptMRG~\cite{jin2024promptmrg} instead predicts finding states and maps them to diagnosis prompts, reaching CE F1 0.476 on MIMIC-CXR without region annotations. Recent methods further incorporate disease-aware representations and adaptive conditioning. Wu et al.~\cite{wu2026diseaseaware} fuse disease-aware semantic tokens with visual features, while Zhao and Liu~\cite{zhao2026multipath} integrate medical language priors and semantic-state guidance in the decoder. RegionGate-CT~\cite{rahman2026regiongate} introduces adaptive region-specific gating for CT report generation but relies on explicit anatomical region decomposition.

In PromptMRG, finding-state prompts guide report generation, but classifier outputs do not modify the visual patch representations used as decoder cross-attention memory. A global channel gate can provide such conditioning, but applying the same gate at every location prevents spatially varying modulation. We address this limitation with a Position-Aware Gate (PAG) that reuses predicted finding states. Each state embedding is combined with a learned finding-identity embedding so that findings sharing the same state remain distinguishable. These finding-specific representations attend to visual patches, and the resulting context produces a scalar gate for each spatial location. PAG therefore conditions the decoder's visual memory using spatially varying signals without requiring region-level annotations.

We also examine the auxiliary alignment objective in the Global+Align reference configuration. Its instance-level InfoNCE loss rapidly approaches zero in our runs and treats every other study as a negative even when studies share positive findings. We therefore propose a Decoder-level Supervised Contrastive Loss (DSCL), adapting supervised contrastive learning~\cite{khosla2020supcon} to the final decoder hidden states. DSCL treats studies sharing at least one ground-truth positive finding as positives, encouraging decoder representations to reflect shared clinical findings rather than instance identity alone.

The principal contributions are:
\begin{itemize}
    \item We propose PAG, which reuses predicted finding states to compute finding-to-patch attention and a scalar gate for each spatial location. PAG provides finding-conditioned spatial modulation without region-level annotations or an auxiliary detection module.
    \item We introduce DSCL, which applies supervised contrastive learning to the final decoder hidden states. Positive pairs are defined by shared ground-truth positive findings, providing finding-aware supervision instead of instance-level alignment.
\end{itemize}

\begin{figure*}[!t]
\centering
\includegraphics[width=\textwidth]{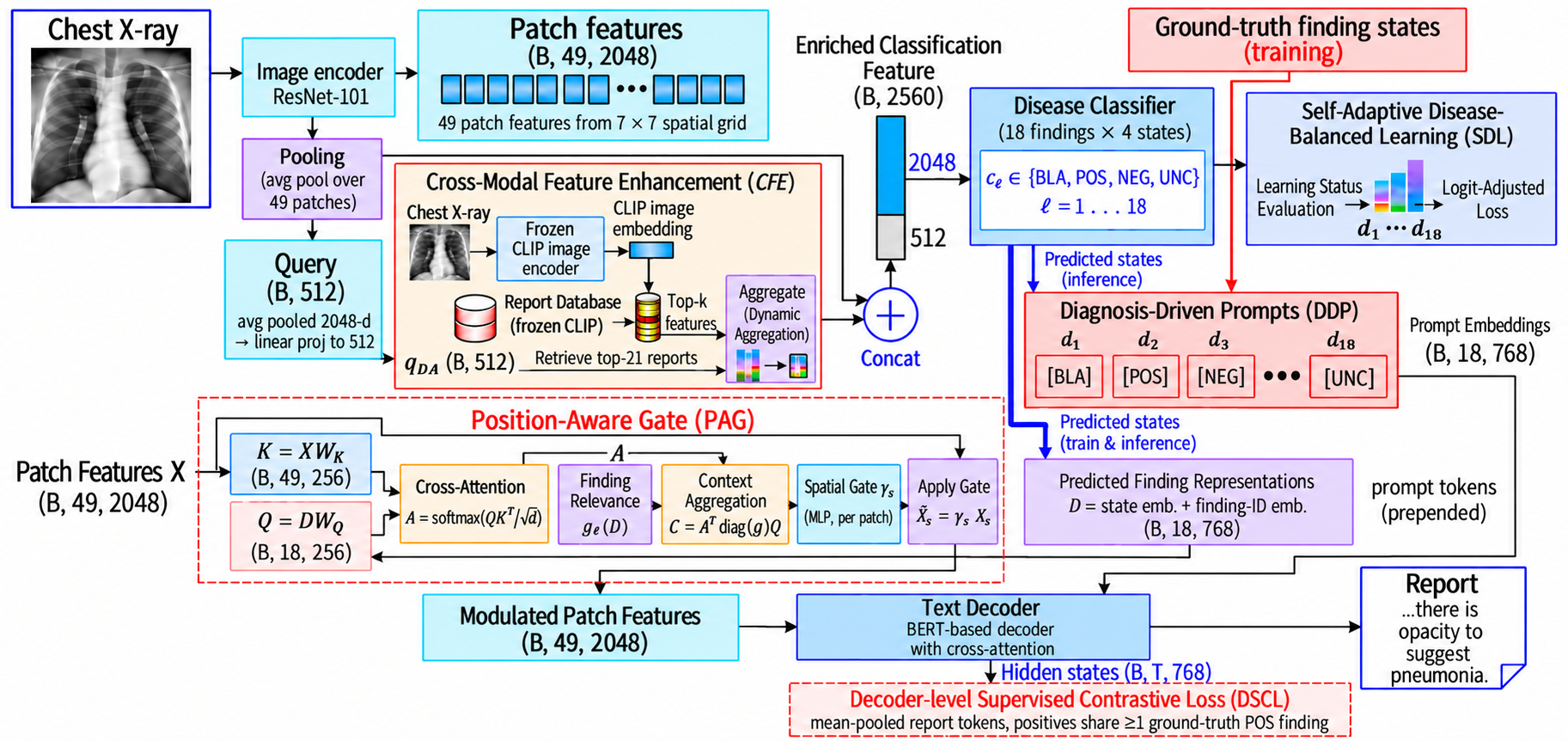}
\caption{Proposed framework. During training, ground-truth finding states provide diagnosis-driven prompt tokens, while classifier-predicted states drive PAG. At inference, classifier predictions serve both roles. PAG fuses state and finding-identity embeddings, attends to $S$ visual patches, and produces a spatial gate $\gamma_s$ for each location. The gated patches serve as decoder cross-attention memory. DSCL supervises pooled decoder representations using positive pairs defined by shared ground-truth positive findings.}
\label{fig:architecture}
\end{figure*}

\section{Proposed Method}
\label{sec:method}

Our framework builds on PromptMRG~\cite{jin2024promptmrg}, as shown in Fig.~\ref{fig:architecture}. We retain its image encoder, cross-modal feature enhancement (CFE), disease classifier, self-adaptive disease-balanced learning (SDL), diagnosis-driven prompts (DDP), and BERT-based decoder. An ImageNet-pretrained ResNet-101 encodes the input into a $7\times7$ feature map ($S=49$), producing patch features $X\in\mathbb{R}^{S\times2048}$. Spatial average pooling gives a global visual feature $\bar{x}_v\in\mathbb{R}^{2048}$. CFE uses a frozen MIMIC-pretrained CLIP model~\cite{endo2021cxrrepair} to retrieve the $k=21$ most similar training reports. Their features are refined by self-attention and fused through cross-attention using a $512$-D projection of $\bar{x}_v$ as query. The resulting retrieval feature $x_r\in\mathbb{R}^{512}$ is concatenated with $\bar{x}_v$ to form the enriched classification representation:
\begin{equation*}
    \bar{x} = \bar{x}_v \oplus x_r
    \in \mathbb{R}^{2560},
\end{equation*}
where $\oplus$ denotes concatenation.

For each of $L=18$ radiographic findings, including 14 CheXbert
categories plus Aorta, Bone/Spine, Hemidiaphragm, and Lung Volume,
the classifier predicts blank (BLA), positive (POS), negative (NEG),
or uncertain (UNC). SDL adjusts the classification logits according to each finding's learning status. During training, DDP maps the 18 ground-truth finding states to special tokens prepended to the BERT-base decoder ($d_t=768$). At inference, these tokens come from classifier predictions. Prompt positions are excluded from the LM loss.

We add two components to address these limitations. PAG replaces
spatially uniform global gating with finding-conditioned, location-dependent
modulation of visual patch features using classifier-predicted finding states.
DSCL replaces instance-level alignment with finding-aware supervision based
on shared positive findings. For controlled comparison, Global+Align
augments PromptMRG with a global channel gate and a symmetric instance-level
InfoNCE objective that encourages agreement between diagnostic conditioning
and the pooled decoder representation. Global+Align serves only as an
experimental reference.

\subsection{Position-Aware Gate (PAG)}
\label{ssec:pag}

In Global+Align, the global channel gate is computed from the enriched classification representation:
\begin{equation*}
\gamma=
\sigma\!\left(W_2\,\mathrm{ReLU}(W_1\bar{x})\right)
\in(0,1)^{2048},
\qquad
\tilde{x}_s=\gamma\odot x_s .
\end{equation*}
Because the same $\gamma$ is applied to every spatial position, it can reweight channels but cannot adapt to individual image locations.

PAG instead computes a location-dependent scalar gate. Let $c_\ell$
denote the predicted state of finding $\ell$. PAG reuses the decoder
state-token embedding $E[c_\ell]\in\mathbb{R}^{d_t}$ used by DDP.
Since different findings can share the same predicted state, a learned
finding-identity embedding $P_\ell$ preserves their identities:
\begin{equation*}
D_\ell=E[c_\ell]+P_\ell,
\qquad
D\in\mathbb{R}^{L\times d_t}.
\end{equation*}
Finding and patch representations are projected into a shared attention space with $d=256$:
\begin{equation*}
Q=DW_Q\in\mathbb{R}^{L\times d},
\qquad
K=XW_K\in\mathbb{R}^{S\times d},
\end{equation*}
followed by finding-to-patch attention:
\begin{equation*}
A=
\mathrm{softmax}\!\left(\frac{QK^\top}{\sqrt{d}}\right)
\in\mathbb{R}^{L\times S},
\end{equation*}
where softmax is applied over the $S$ patches for each finding.

Because each row of $A$ sums to one, it captures relative spatial association but not each finding's overall relevance. We therefore learn a finding-level relevance score and form patch context vectors:
\begin{equation*}
g_\ell=\sigma(w_g^\top D_\ell+b_g),
\qquad
C=A^\top\mathrm{diag}(g)Q
\in\mathbb{R}^{S\times d},
\end{equation*}
where $g=[g_1,\ldots,g_L]$. The relevance term allows findings to contribute unequally despite the normalized attention rows. A shared two-layer MLP then produces one scalar gate per spatial location:
\[
\gamma_s=\sigma\!\left(\mathrm{MLP}(C_s)\right),
\qquad
\tilde{x}_s=\gamma_s x_s,
\]
with dimensions $256\!\rightarrow\!128\!\rightarrow\!1$. Thus, $\tilde{X}\in\mathbb{R}^{S\times2048}$ replaces $X$ as decoder cross-attention memory, allowing spatially varying modulation while preserving the original feature dimensionality.

The final MLP layer is initialized with zero weights and bias $2$, giving $\gamma_s=\sigma(2)\approx0.88$ initially. PAG uses classifier-predicted finding states during both training and inference and receives no ground-truth states. This keeps the gating pathway consistent with inference and avoids direct ground-truth leakage into PAG. Ground-truth findings condition the DDP prompt, while predicted findings condition the visual memory. PAG adds approximately $0.77$\,M parameters and computes an $18\times49$ finding-to-patch attention map per image.

\subsection{Decoder-level Supervised Contrastive Loss}
\label{ssec:dscl}

Global+Align uses a symmetric instance-level InfoNCE objective. In our runs, this loss approaches zero within approximately 500 optimization steps. More importantly, instance-level alignment treats all other studies as negatives even when they share positive findings. DSCL instead applies supervised contrastive learning~\cite{khosla2020supcon} to the final decoder hidden states using clinical findings to define sample relationships.

Let $H\in\mathbb{R}^{B\times T\times d_t}$ denote the final-layer decoder hidden states for a minibatch of $B$ studies. After removing diagnosis-prompt tokens and padding, the remaining report-token states are mean-pooled and $\ell_2$-normalized:
\begin{equation*}
h_i=\frac{1}{|\mathcal{T}_i|}\sum_{t\in\mathcal{T}_i}H_{i,t},
\qquad
z_i=\frac{h_i}{\|h_i\|_2},
\end{equation*}
where $\mathcal{T}_i$ contains the valid report-token positions for study $i$.

Let $y_{i\ell}$ denote the ground-truth state of finding $\ell$ for study $i$. Two studies form a positive pair if they share at least one ground-truth positive finding:
\begin{equation*}
P(i)=
\left\{
j\neq i:
\exists\,\ell,\;
y_{i\ell}=y_{j\ell}=\mathrm{POS}
\right\}.
\end{equation*}
For valid anchors $\mathcal{A}=\{i:|P(i)|>0\}$, DSCL is
\begin{equation*}
\mathcal{L}_{\mathrm{SC}}
=
\frac{1}{|\mathcal{A}|}
\sum_{i\in\mathcal{A}}
\frac{-1}{|P(i)|}
\sum_{j\in P(i)}
\log
\frac{\exp(z_i^\top z_j/\tau)}
{\sum_{k\neq i}\exp(z_i^\top z_k/\tau)},
\end{equation*}
with $\tau=0.07$. If no valid positive pair exists in a minibatch, we set $\mathcal{L}_{\mathrm{SC}}=0$. DSCL therefore changes the contrastive relation from instance identity to shared positive findings, encouraging clinically related reports to have similar decoder representations. Ground-truth findings are used only to construct training-time pairs, so DSCL adds no inference-time computation.

The complete objective is
\begin{equation*}
\mathcal{L}
=
\mathcal{L}_{\mathrm{LM}}
+
\lambda_{\mathrm{cls}}\mathcal{L}_{\mathrm{cls}}
+
\lambda_{\mathrm{sc}}\mathcal{L}_{\mathrm{SC}},
\end{equation*}
where $\mathcal{L}_{\mathrm{LM}}$ is the generation loss and $\mathcal{L}_{\mathrm{cls}}$ is the SDL-adjusted classification loss inherited from PromptMRG. We use $\lambda_{\mathrm{cls}}=4.0$ and $\lambda_{\mathrm{sc}}=0.3$. During training, the image encoder, CFE, disease classifier, PAG, and decoder are optimized jointly while CLIP remains frozen. At inference, predicted finding states condition both DDP and PAG, and the decoder generates the report from the diagnosis-driven prompts and modulated visual features $\tilde{X}$.

\section{Experimental Results}

We evaluate on MIMIC-CXR~\cite{johnson2019mimiccxr} and IU X-Ray~\cite{demnerfushman2016iu}. Following PromptMRG~\cite{jin2024promptmrg}, the preprocessed MIMIC-CXR split contains 270{,}790/2{,}130/3{,}858 training/validation/test instances, with images resized and center-cropped to $224\times224$. For IU X-Ray, we use the same 4{,}168-image cross-dataset evaluation set and protocol as PromptMRG. The MIMIC-CXR-trained model is evaluated directly without IU X-Ray fine-tuning. We report CE precision, recall, and F1 over 14 CheXpert categories~\cite{irvin2019chexpert} using CheXbert~\cite{smit2020chexbert}, together with BLEU-1, BLEU-4, and ROUGE-L. CE F1 is the primary metric.

All controlled configurations are trained for 10 epochs with AdamW
using $\beta=(0.9,0.999)$, weight decay 0.05, and per-GPU batch size
16 on 4 NVIDIA RTX A5000 GPUs. The learning rate uses a 2,000-step
warm-up to $5\times10^{-5}$ followed by cosine decay to
$5\times10^{-6}$. Gradients are clipped to a norm of 0.1. PAG uses
$d=256$, while DSCL uses $\tau=0.07$ and
$\lambda_{\mathrm{sc}}=0.3$. DSCL positives and negatives are
constructed within each per-GPU minibatch. Decoding uses beam size 3,
and the checkpoint with the highest validation CE F1 is selected. All
controlled experiments use the same data split, training, decoding,
and random-seed settings.

\begin{table}[t]
\centering
\caption{Results on MIMIC-CXR. $\dagger$ indicates values taken from
PromptMRG. $\#$ indicates a different split and input resolution
$>224\times224$. Bold indicates the best comparable result shown.}
\label{tab:main}
\small
\setlength{\tabcolsep}{1.25pt}
\renewcommand{\arraystretch}{0.96}
\begin{tabular}{@{}lcccccc@{}}
\toprule
Method & CE-P & CE-R & CE-F1 & B-1 & B-4 & R-L \\
\midrule
R2Gen$^\dagger$~\cite{chen2020r2gen}
& 0.333 & 0.273 & 0.276 & 0.353 & 0.103 & 0.277 \\
METrans.$^\dagger$~\cite{wang2023metransformer}
& 0.364 & 0.309 & 0.311 & 0.386 & \textbf{0.124} & \textbf{0.291} \\
KiUT$^\dagger$~\cite{huang2023kiut}
& 0.371 & 0.318 & 0.321 & 0.393 & 0.113 & 0.285 \\
DCL$^\dagger$~\cite{li2023dcl}
& 0.471 & 0.352 & 0.373 & -- & 0.109 & 0.284 \\
RGRG$^{\dagger\#}$~\cite{tanida2023rgrg}
& 0.461 & 0.475 & 0.447 & 0.373 & 0.126 & 0.264 \\
PromptMRG$^\dagger$~\cite{jin2024promptmrg}
& 0.501 & 0.509 & 0.476 & 0.398 & 0.112 & 0.268 \\
\midrule
Global+Align
& 0.504 & 0.511 & 0.484 & 0.414 & 0.113 & 0.263 \\
PAG+DSCL
& \textbf{0.527} & \textbf{0.542} & \textbf{0.502}
& \textbf{0.422} & 0.116 & 0.284 \\
\bottomrule
\end{tabular}
\end{table}

\begin{table}[t]
\centering
\caption{Cross-dataset evaluation on IU X-Ray following PromptMRG.
$\dagger$ indicates values taken from PromptMRG. $*$ indicates input
resolution $>224\times224$. Bold indicates the best comparable result shown.}
\label{tab:iu}
\small
\setlength{\tabcolsep}{1.35pt}
\renewcommand{\arraystretch}{0.96}
\begin{tabular}{@{}lcccccc@{}}
\toprule
Method & CE-P & CE-R & CE-F1 & B-1 & B-4 & R-L \\
\midrule
R2Gen$^\dagger$~\cite{chen2020r2gen}
& 0.141 & 0.136 & 0.136 & 0.325 & 0.059 & 0.253 \\
CVT2Dis.$^{*\dagger}$~\cite{nicolson2022cvt2dis}
& 0.174 & 0.172 & 0.168 & 0.383 & 0.082 & 0.277 \\
DCL$^\dagger$~\cite{li2023dcl}
& 0.168 & 0.167 & 0.162 & 0.354 & 0.074 & 0.267 \\
RGRG$^{*\dagger}$~\cite{tanida2023rgrg}
& 0.183 & 0.187 & 0.180 & 0.266 & 0.063 & 0.180 \\
PromptMRG$^\dagger$~\cite{jin2024promptmrg}
& 0.213 & 0.229 & 0.211 & 0.401 & \textbf{0.098} & 0.281 \\
\midrule
PAG+DSCL
& \textbf{0.231} & \textbf{0.247} & \textbf{0.226}
& \textbf{0.431} & 0.093 & \textbf{0.302} \\
\bottomrule
\end{tabular}
\end{table}

On MIMIC-CXR, PAG+DSCL improves CE F1 from 0.484 for Global+Align to 0.502, with CE precision and recall increasing to 0.527 and 0.542. PromptMRG reports 0.476 CE F1 under the same preprocessing. Generation metrics also improve over Global+Align, reaching 0.422/0.116/0.284 for BLEU-1, BLEU-4, and ROUGE-L.

Without IU X-Ray fine-tuning, PAG+DSCL achieves CE precision, recall, and F1 of 0.231, 0.247, and 0.226. Under the same PromptMRG protocol, the reported values are 0.213, 0.229, and 0.211. BLEU-1 and ROUGE-L increase to 0.431 and 0.302, while BLEU-4 decreases slightly from 0.098 to 0.093. Across both datasets, CE and NLG metrics do not improve uniformly, which is consistent with PAG and DSCL operating on finding-level clinical information while clinically similar reports may differ lexically.

Table~\ref{tab:ablation} separates the contributions of PAG and DSCL. Replacing the global gate with PAG raises CE F1 from 0.484 to 0.491, while replacing instance-level alignment with DSCL reaches 0.495. Combining both yields 0.502. The parameter-matched global gate reaches 0.483, suggesting that PAG's improvement is not explained solely by additional parameters.

The reference alignment loss approaches zero within approximately
500 optimization steps. For an anchor with $|P(i)|$ positives, DSCL
has a lower bound of $\log |P(i)|$, so the two loss magnitudes are not
directly comparable. More importantly, DSCL treats studies sharing
positive findings as positives rather than treating all other studies
as negatives. PAG adds approximately $0.77$\,M parameters and computes
an $18\times49$ attention map per image, while DSCL adds no
inference-time computation.

\begin{table}[t]
\centering
\caption{Component ablation on MIMIC-CXR. PM denotes a parameter-matched
global gate with hidden dimension 166. $\Delta$ is relative to
Global+Align.}
\label{tab:ablation}
\small
\setlength{\tabcolsep}{1.30pt}
\renewcommand{\arraystretch}{0.96}
\begin{tabular}{@{}llcccc@{}}
\toprule
\# & Configuration & CE-P & CE-R & CE-F1 & $\Delta$ \\
\midrule
1 & Global+Align
& 0.504 & 0.511 & 0.484 & 0.000 \\
2 & PAG+Align
& 0.511 & 0.531 & 0.491 & +0.007 \\
3 & Global+DSCL
& 0.514 & 0.536 & 0.495 & +0.011 \\
4 & PAG+DSCL
& \textbf{0.527} & \textbf{0.542} & \textbf{0.502} & +0.018 \\
\midrule
5 & Global-PM+Align
& 0.502 & 0.522 & 0.483 & -0.001 \\
\bottomrule
\end{tabular}
\end{table}

\section{Conclusion}
\label{sec:conclusion}

We propose a Position-Aware Gate (PAG) and Decoder-level Supervised Contrastive Loss (DSCL) for diagnosis-conditioned radiology report generation. PAG spatially modulates visual patches using predicted finding states, while DSCL structures decoder representations using shared positive findings. On MIMIC-CXR, CE F1 increases from 0.484 to 0.491 with PAG, 0.495 with DSCL, and 0.502 when combined. A parameter-matched global gate reaches 0.483. Without IU X-Ray fine-tuning, PAG+DSCL achieves 0.226 CE F1 compared with 0.211 reported by PromptMRG. Limitations include evaluation on two datasets, reliance on CheXbert-based CE metrics, a fixed $7\times7$ visual grid, and a permissive DSCL positive-pair definition. Future work will explore broader datasets, adaptive spatial representations, radiologist evaluation, and stricter clinical similarity criteria.


\section{Compliance with Ethical Standards}
This study used de-identified MIMIC-CXR and IU X-Ray data. MIMIC-CXR was accessed via PhysioNet under its credentialed data use agreement.

\section{Acknowledgements}
This work was supported by the Department of Computer Science,
Texas State University. The authors declare no relevant conflicts
of interest. ChatGPT (OpenAI) was used for language editing.
All technical content was verified by the authors.

\bibliographystyle{IEEEbib}
\bibliography{strings,refs}

\end{document}